\documentclass[conference]{IEEEtran}
\usepackage{times}
\usepackage[numbers]{natbib}
\usepackage{multicol}
\usepackage[bookmarks=true]{hyperref}
\usepackage{graphicx}
\usepackage{amsmath}
\usepackage{amsfonts}
\usepackage{multirow}
\usepackage{threeparttable}
\usepackage{booktabs}
\usepackage[caption=false,font=small,labelfont=rm,textfont=rm]{subfig}
\usepackage{amsfonts}
\usepackage{amssymb}
\usepackage{xcolor}
\usepackage{algorithmic}
\usepackage[ruled,vlined]{algorithm2e}

\definecolor{light-gray}{gray}{0.95}

\renewcommand\fbox{\fcolorbox{light-gray}{bg-gray}}
\definecolor{dark-gray}{HTML}{A9A9A9} 
\definecolor{light-gray}{HTML}{b7b7b7} 
\definecolor{bg-gray}{HTML}{F8F8F8} 
	
\definecolor{green}{HTML}{6aa84f} 
\definecolor{highlight}{HTML}{cfe2f3} 
\definecolor{blue}{HTML}{0000ff} 
\definecolor{brown}{HTML}{9F8C76}
\definecolor{pink}{HTML}{D88782}
\definecolor{light-blue}{HTML}{4a86e8}
\definecolor{dark-yellow}{HTML}{F6AE2D}
\definecolor{light-pink}{HTML}{ffcccc}
\definecolor{light-yellow}{HTML}{FFE5B4}

\begin{document}
\title{SViP: Sequencing Bimanual Visuomotor Policies with Object-Centric Motion Primitives}

\author{\authorblockN{
Yizhou Chen$^{1,2}$,
Hang Xu$^{1,2}$,
Dongjie Yu$^{1}$,
Zeqing Zhang$^{1}$, 
Yi Ren$^{3}$ and
Jia Pan$^{1}$,
}
\smallskip
\authorblockA{
$^{1}$ The University of Hong Kong
$^{2}$ Centre for Transformative Garment Production (InnoHK)
$^{3}$ Huawei Technologies Co., Ltd \\
}
}

\maketitle

\begin{abstract}

Imitation learning (IL), particularly when leveraging high-dimensional visual inputs for policy training, has proven intuitive and effective in complex bimanual manipulation tasks. Nonetheless, the generalization capability of visuomotor policies remains limited, especially when small demonstration datasets are available. 
Accumulated errors in visuomotor policies significantly hinder their ability to complete long-horizon tasks. To address these limitations, we propose \textit{SViP}, a framework that seamlessly integrates visuomotor policies into task and motion planning (TAMP).  SViP partitions human demonstrations into bimanual and unimanual operations using a semantic scene graph monitor. Continuous decision variables from the key scene graph are employed to train a switching condition generator. This generator produces parameterized scripted primitives that ensure reliable performance even when encountering out-of-the-distribution observations.
Using only 20 real-world demonstrations, we show that SViP enables visuomotor policies to generalize across out-of-distribution initial conditions without requiring object pose estimators. For previously unseen tasks, SViP automatically discovers effective solutions to achieve the goal, leveraging constraint modeling in TAMP formulism. In real-world experiments, SViP outperforms state-of-the-art generative IL methods, indicating wider applicability for more complex tasks. Project website: \href{https://sites.google.com/view/svip-bimanual}{https://sites.google.com/view/svip-bimanual}.
\end{abstract}

\IEEEpeerreviewmaketitle

\section{Introduction}

Bimanual manipulation poses significant challenges in tasks that involve multiple stages and require precise hand-hand coordination \cite{krebs2022bimanual,yu2024bikc, chen2019combined}. Traditional motion planning often relies on accurate modeling of the objects being manipulated, which may not be feasible in real-world applications. By contrast, Learning-from-Demonstration (LfD) circumvents these limitations by enabling robots to learn actuation commands directly from human demonstrations, thus eliminating the need for intricate object modeling. 

Recent developments in IL methods, designing the visuomotor policies as denoising diffusion probabilistic models (DDPMs) \cite{chi2023diffusion, ke20243d, hu2024stem, ze20243d}, have significantly enhanced their modeling capabilities for long-horizon and multimodal manipulations. Nevertheless, these methods still exhibit a high vulnerability to failures when faced with out-of-distribution (OOD) observations.
For instance, the gripper might fail to pick up an object if the object is displaced by only a few centimeters from its demonstrated position. Moreover, even if the picking is successful, subsequent manipulation can still fail due to an unseen grasp gesture. 
To learn a more reliable policy in challenging scenarios, recent works \cite{black2024pi0,zhao2024aloha, liu2024rdt} leverage large-scale datasets that encompass a wide distribution. However, the substantial hardware requirements (e.g., computational power and storage capacity) and the intensive human effort (e.g., demonstration data collection) often render these approaches impractical for real-world industrial applications. 
Another issue is the constrained manipulability of teleoperation hardware systems. These systems employing direct joint mapping (e.g., ALOHA\cite{zhao2023learning} and GELLO \cite{wu2024gello}) are widely used for collecting fine-grained manipulation data. However, human experts using such systems often struggle to reach a large portion of configurations within diverse tabletop setups -- configurations that robotic arms can readily access..

\begin{figure}
    \centering
    \includegraphics[width=1\linewidth]{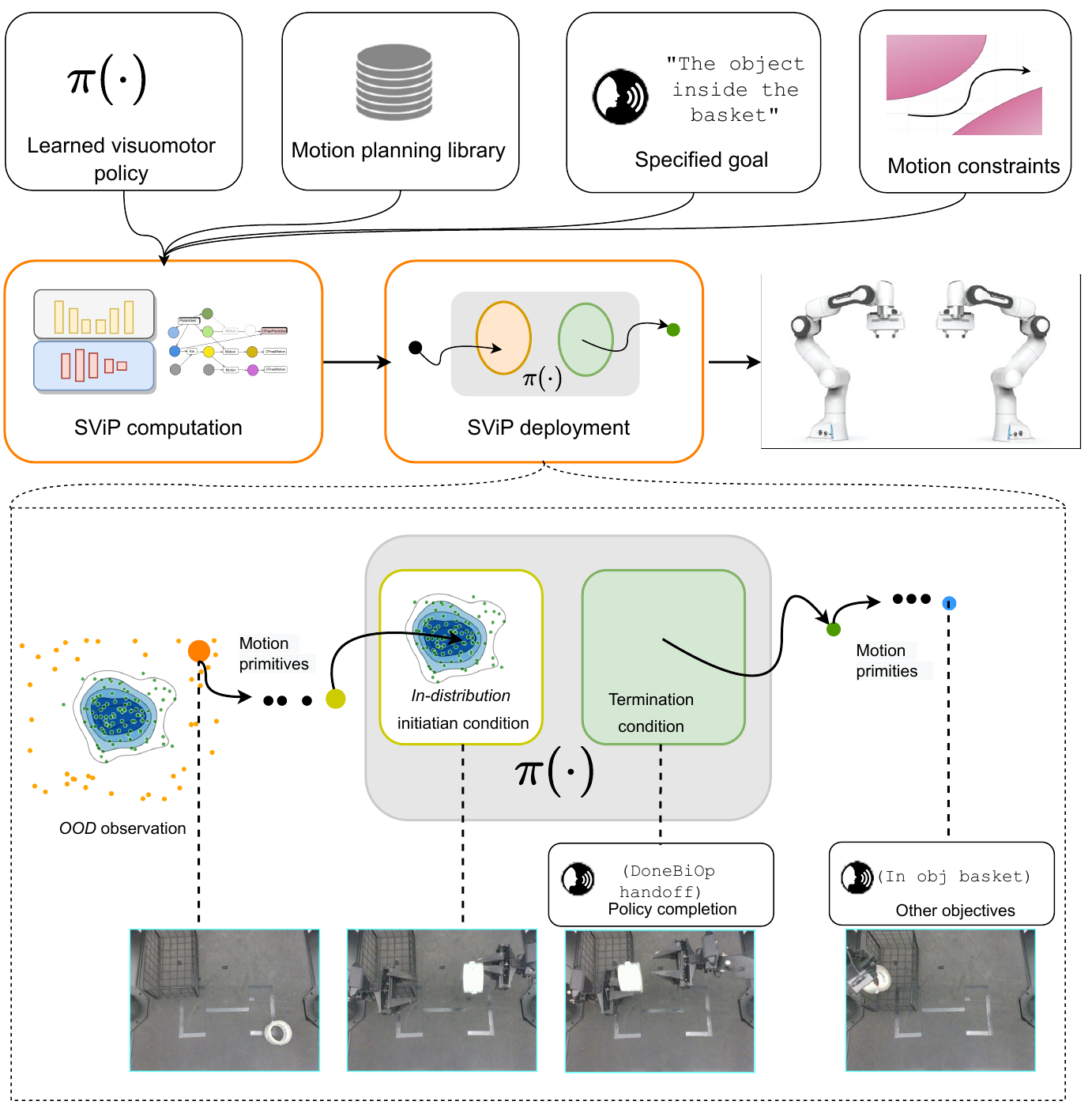}
    \caption{Provided with motion planning primitives and trained bimanual visuomotor policies, \textbf{SViP} utilizes task and motion planning to compute a plan skeleton that integrates both learned and planned operations. When encountering observations absent from the training data, \textbf{SViP} leverages planned operations to transition into a state within the training distribution, enabling the initiation of the learned bimanual visuomotor policy. Additionally, \textbf{SViP} facilitates the completion of customized goals while adhering to specified motion constraints.}
    \label{fig:ood-2-id}
    \vspace{-0.2in}
\end{figure}

It is noteworthy that classical motion planning approaches do not rely on demonstrations, and they avoid inherent limitations like poor generalization or accumulated error.
Therefore, we aim to integrate the strengths of both learned visuomotor policy and scripted motion planning. Specifically, we propose \textbf{S}eqeuncing \textbf{Vi}suomotor \textbf{P}olicy (\textbf{SViP}) that utilizes task and motion planning (TAMP) to stitch learned visuomotor policies and object-centric motion primitives into a cohesive sequence to accomplish tasks, as pictured in Fig. \ref{fig:ood-2-id}. Instead of learning a fixed action sequence, SViP allows robots to perform complex tasks that involve multiple branches and diverse operations.
This work is inspired by \citet{mandlekar2023human}, which enables scripted motion, visuomotor policy, and human teleoperation to work jointly in an interleaved fashion. However, it requires a predefined and hardcoded switching mechanism between scripted and learned operations. Its successor \citep{garrett2024skillmimicgen} combines motion planning with demonstration to generate training data under different environmental conditions. 
A common limitation of \citet{mandlekar2023human} and \citet{garrett2024skillmimicgen} is their reliance on 6-DoF object poses, which necessitates the use of pose estimators. Unfortunately, these estimators require extra tedious data collection and long training time, and their performance is often suboptimal for objects exhibiting symmetry \cite{liu2022gen6d, fan2024pope, wen2024foundationpose}. 
In contrast to the aforementioned approaches \cite{mandlekar2023human,garrett2024skillmimicgen}, SViP directly learns a mapping from point cloud observations to switching conditions, ensuring that visuomotor policies are initiated based on observations similar to those encountered in demonstrations (Fig. \ref{fig:ood-2-id} bottom). {Furthurmore, unlike prior methods such as NOD-TAMP \cite{cheng2023nod} and PSL \cite{dalal2024plan} that stitch fixed sequences of unimanual skills, SViP offers broader applicability and greater composition flexibility.}
The contributions of the proposed SViP are summarized as follows: 
\begin{itemize}
    \item We propose a compositional system that can decompose unimanual or bimanual skills from demonstrations and reorganize them to achieve novel goals;
    \item We design switching condition generators and feasibility validators for visuomotor policies,  facilitating their  seamless sequencing with motion primitives while complying with motion constraints;
    \item We conduct comprehensive experiments in both simulation and real world, demonstrating that SViP can complete long-horizon manipulation in challenging setups.
\end{itemize}

\section{Preliminaries}

\subsection{Scene Graph} \label{sec:preliminary_sg}
A scene graph can be defined as a graph $G = \{V,E, L\}, $ where $V$ is the set of entities as nodes,  $E \subseteq V \times V$ is the set of edges, and $L: E \rightarrow \mathbb{P}$ is a function that maps an edge to a predicate which represents a fixed transform or a kinematic link between entities. We consider three types of entities in a bimanual tabletop manipulation setting: robot gripper nodes $H$, region nodes $R$, and object nodes $O$. An edge $e \in E$ in a scene graph captures the contact mode in manipulation, which can in turn be converted to a predicate. For example, an edge $ e \in \{  \langle h, o \rangle \vert h \in H, o \in O \}$ corresponds to $\mathsf{AtGrasp}(h, o, g)$, where  $g $ is the relative gripper pose to the object frame;  an edge $e \in \{ \langle \rho_0, h \rangle \vert h \in H \}$ corresponds to $\mathsf{AtConf}( h, q)$, where $\rho_0$ is the table and $q$ is the joint configurations;  an edge $e \in \{ \langle \rho, o \rangle \vert \rho \in R, o \in O \}$ corresponds to $\mathsf{AtRelativePose}(\rho, o, p)$, where $p$ is the relative pose of  $o$ in the frame of $\rho$.   
In the following context, we use  $q$ to denote joint angles, and  $g_{o,h}$ to denote the contacting pose of gripper $h$ on object $o$. The subscripts $l$ or $r$ are used to specify the related arm, i.e., left or right arm.  For ease of illustration, we simplify $\mathsf{AtRelativePose}(\rho_0, o, p)$ to $\mathsf{AtPose}(o, p)$, since the world frame is set to the table frame. Similarly, we abbreviate $g_{o,h}$ as $g_o$, as each object is contacted by at most one gripper in an object-centric unimanual operation. We denote $\tau_{o}$ as the object-centric trajectory when breaking or establishing contact with the object $o$; the contacting pose $g_o$ is trivially obtained from $\tau_{o}$ when the gripper is closest to the object.

Let $Gr(s_t)$  be a mapping from system state $s_t$ to a scene graph $G$. For a demonstration ${D}$ containing $T$ timesteps, an \textit{event-driven scene graph sequence} can be sketched according to the contact mode changes:
$$
\mathbf{G}(D) = Gr(s_0)Gr(s_{t_1})Gr(s_{t_1+t_2}) \cdots Gr(s_{\sum t_i})
$$
where ${\sum t_i}<T$ and $Gr(s_{t_i}) \neq Gr(s_{t_{i+1}})$. 

\section{Methodology} \label{sec method}

\subsection{Symbolic Descriptions for Bimannual Visuomotor Policies} \label{sec:components}

We define an abstracted bimanual skill (referred to as \textit{skill} below) as a tuple $ \mathfrak{a} = \langle G_\mathrm{mid}, G_\mathrm{pre}, G_\mathrm{eff}, \pi, \varphi, \xi \rangle$, where $G_\mathrm{mid}$ is the scene graph during the contact-rich segment of bimanual manipulation, $G_\mathrm{pre}$ is the scene graph preceding $G_\mathrm{mid}$, $G_\mathrm{eff}$ is the scene graph immediately following $G_\mathrm{mid}$, $\pi$ is the learned visuomotor policy, $\varphi$ is the switching condition generator, and $\xi$ is the feasibility validator.
The intuition behind scene-graph representations derives from the observation that a demonstrated skill can be decomposed into several sub-skills according to contact modes between robot hands and manipulated objects. The contact-rich segments pose challenges for classical motion planning or reward function specification, yet are tractable for a learned visuomotor policy $\pi$. On the other hand, the segments with lower behaviour complexity (e.g., pick, place,  drop, transit, and transfer) are better addressed by classical motion planning.

\begin{figure}[t]
    \centering
    \includegraphics[width=\linewidth]{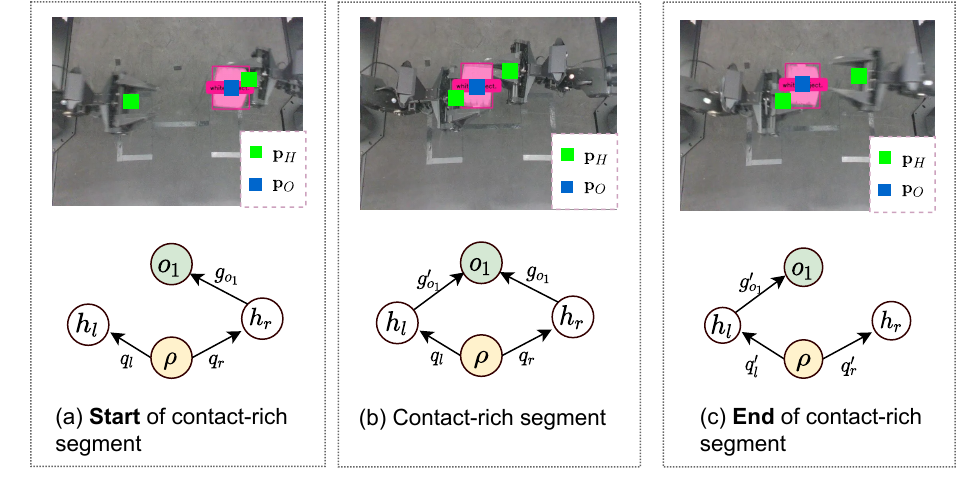}
    \vspace{-0.25in}
    \caption{(a) and (c) show the \textit{start} and the \textit{end} of the contact-rich part, while (b) shows the contact-rich part of the \textit{Object Handoff} operation. As marked in the overhead camera images, $p_H$ and $p_O$ are the center points of robot grippers and objects. The corresponding scene graphs $G_\mathrm{pre}, G_\mathrm{mid}, G_\mathrm{eff}$ are sketched below, with continuous variables labeled on each edge. Here, $\rho$, $h_l$, $h_r$, $o_1$ denote the table, the left robot, the right robot, and the manipulated objects, respectively. The edges within each scene graph are labeled with continuous decision variables.  }
    \label{fig:transfer_sg}
    \vspace{-0.1in}
\end{figure}

We develop a perception pipeline that segments a demonstration into a sequence of scene graphs. These graphs represent subgoals that can be achieved by visuomotor policies or object-centric motion primitives.
Via a video segmentation tool \cite{ren2024grounded},  the monitor obtains the positions of all objects $p_{O}$ and the positions of all robot end-effectors $p_{H}$ (Fig. \ref{fig:transfer_sg} top). An event-driven scene graph sequence can thus be constructed as $\mathbf{G}(D)$, as sketched in Fig. \ref{fig:transfer_sg} bottom. 
{Specifically, the scene graph $G_\mathrm{mid}$  represents the segment where there is contact between objects or coordination between grippers.  On the other hand, scene graphs $G^\mathfrak{a}_\mathrm{pre}$ and $G^\mathfrak{a}_\mathrm{eff}$ contains vital geometric information before and after the contact-rich segment. They are further parsed into   PDDLStream\cite{garrett2020pddlstream}   format, which provides highly abstracted descriptions of skills for symbolic planning. }

Consider a bimanual \textit{Object Handoff} task in Fig.~\ref{fig:transfer_sg} as an example, where an object $o_1$ is to be handed off from the right gripper $h_r$ to the left gripper $h_l$.  Let  $\mathsf{Pr}(G): {G} \rightarrow \mathbb{P}$ be a function that converts a scene graph into a set of predicates with grounded symbolic variables. Thus, the geometric information from $G^\mathfrak{a}_\mathrm{pre}$ is converted to the predicates  $ \mathsf{PreGeom}(o_1, h_l, h_r, q_l, q_r, g_{o_1}) =  \mathsf{Pr}(G^\mathfrak{a}_\mathrm{pre}) =\mathsf{AtGrasp} (h_r ,o_1 ,g_{o_1}) \wedge \mathsf{AtConf} (h_l ,q_l)  \wedge \mathsf{AtConf} (h_r ,q_r)$. Similarly, the contact-related predicates in $\mathrm{eff}(\mathfrak{a})$ is extracted by comparing $\mathsf{Pr}(G^\mathfrak{a}_\mathrm{eff}) $ to $\mathsf{Pr}(G^\mathfrak{a}_\mathrm{pre}) $, resulting in $ \mathsf{EffGeom}(o_1, h_l, h_r,q'_l, q'_r, g_{o_1}, g'_{o_1}) = \mathsf{AtGrasp} (h_l ,o_1 ,g'_{o_1}) \wedge \neg \mathsf{AtGrasp} (h_r ,o_1 ,g_{o_1}) \wedge \mathsf{AtConf} (h_l ,q'_l)  \wedge \mathsf{AtConf} (h_r ,q'_r)$. Other than $\mathsf{EffGeom}$, the effect contains semantic information about the completion of the bimanual operation, i.e.,  $\mathsf{DoneBiOp}(\mathfrak{a})$. Thus, the description of the bimanual operation $\mathfrak{a}$ is given as:
$$
\begin{aligned}
\mathsf{BiOperation}&(\mathfrak{a}, o_1, h_l, h_r, q_l, q_r, q'_l, q'_r, g_{o_1}, g'_{o_1}) \\
 \mathbf{pre:}&   \mathsf{PreGeom}(o_1, h_l, h_r, q_l, q_r, g_{o_1}) \\
\mathbf{eff:}&   \mathsf{EffGeom}(o_1, h_l, h_r,q'_l, q'_r, g_{o_1}, g'_{o_1}) \wedge \\ & \mathsf{DoneBiOp}(\mathfrak{a}) \\
 \mathbf{con:}&  \mathsf{SafeBiOp}(\mathfrak{a},h_l, h_r, q_l, q_r) 
 \vspace{-1mm}
\end{aligned}
$$

This description can be easily generalized to other learned policies with available geometric information $G_\mathrm{pre}, G_\mathrm{eff}$. If the demonstration ends with a contact-rich operation, geometric predicates in $G_\mathrm{eff}$ are not available, thus the effect should only contain $\mathsf{DoneBiOp}$ as a sign of completion. Otherwise, if geometric predicates are available in the effect, the termination of the policy $\pi$ can be reported by a sub-goal monitor.

\subsection{Diffusion-based Parameter Generator} \label{sec paragen}
Encoding a learned policy into TAMP is non-trivial because there is a significant gap between the formulation of visuomotor policy and the planning. Our idea is similar to that in \citep{mandlekar2023human, garrett2024skillmimicgen}, executing the learned policy and planned trajectories in an interleaved way. The bottleneck of a hybrid learned-scripted system is to develop a suitable switching mechanism. Unlike prior works that only focus on unimanual operations~\cite{mandlekar2023human, garrett2024skillmimicgen, cheng2023league}, we involve the sequencing of complex bimanual operations. To stitch object-centric motion primitives to the initiation and termination of a learned policy, we train networks to predict the switching condition for unimanual and bimanual operations separately.

For object-centric unimanual operations, we learn a trajectory $\tau_o$ in the contact-rich segment.
On the other hand,  bimanual operations are modeled as black-box skills starting and ending at certain configurations $\mathbf{q}$. 
Let $\mathcal{V}$ denote the collection of learnable decision variables $\{\mathbf{q},  \tau_{o_1}, \dots , \tau_{o_M}\}$. Assuming  independent  behavior for each atomic skill, we can factorize the distribution of $\mathcal{V}$ as:
\begin{equation}
\label{eq:maxprob}
    \vspace{-1mm}
    P(\mathcal{V}| C_{o_1}, \dots , C_{o_M} ) = \  P ( \mathbf{q}) \ \cdot\prod^{M}_{i=1} P(\tau_{o_i}|C_{o_i}) 
\end{equation}
where $\mathbf{q} = \langle q_\mathrm{pre}, q_\mathrm{eff} \rangle$ as the joint angle array concatenating the starting and end joint angles, $M$ is the number of object-centric operations, $C_{o_i}$ is the point cloud of the object $o_i$ being contacted. Note that Equation \ref{eq:maxprob} includes one bimanual skill for simplicity, and it can be extended to demonstration sequences containing multiple bimanual skills.

{We formulate the switching condition generators as DDPM due to its powerful distribution fitting capability. According to \citet{salimans2021should}, the connection of score function and probability distribution can be established as $\varepsilon(x,k) = \nabla_x \log (P(x))$, where $k$ be the diffusion step. Thus, Equation \ref{eq:maxprob}  can be rewritten as  a combination of score functions:
\begin{equation}
\label{eq:combinescore}
   \hat{ \varepsilon}( \mathcal{V} | C_{o_1}, \dots , C_{o_M} ) = \varepsilon_{\theta} ( \mathbf{q}, k) + \sum^{M}_{i=1} \varepsilon_{\theta}(\tau_{o_i},k|C_{o_i}) 
\end{equation}
where  $\varepsilon_\theta$ is an approximate score function.  Specifically, an unconditional score function for starting/termination joint poses is sampled from a multi-layer perceptron (MLP), while a score function for object-centric trajectory is sampled from an $SE$(3) equivariant U-Net \cite{yang2024equibot}.}
The $SE$(3) equivariance is achieved by the Vector Neuron Network (VNN) \cite{deng2021vector} architecture, enforcing equivalent distribution of the conditional probability $P(\tau^g_o | C_o)$ and $P(\mathcal{T} \tau^g_o | \mathcal{T} C_o)$ under arbitrary homogeneous transformation $\mathcal{T} \in SE(3)$.  Accordingly, the point cloud feature is extracted by an $SE$(3)-equivariant point cloud encoder $\mathsf{Enc}(\cdot)$, adopted from \citet{lei2023efem}.
The noise prediction function in the $k^{th}$ denoising step can thus be written as $\epsilon_\theta (\mathsf{Enc}(C_o), \mathsf{Vec}(\tau_o) + \varepsilon^k, k) = \varepsilon^k$, where  $\mathsf{Vec}(\cdot)$ extracts the rotation part in an \textit{SE}(3) pose following \citet{zhou2019continuity}. 

{With the generators of switching conditions wrapped in stream functions, we use Adaptive algorithm \cite{garrett2018sampling} to solve the TAMP problem, computing an action sequence composed of visuomotor policies or learned trajectories. }

\subsection{Constraints in Bimanual Visuomotor Policies} \label{sec feasibility}

With the building blocks introduced above,  SViP leverages \textit{focused} TAMP solver \citep{garrett2018sampling} to search for a set of discrete and continuous parameters that satisfy various constraints. However, the calculated action sequence is not guaranteed feasible, as constraints associated with the policy are not well-defined. To ensure that the bimanual visuomotor policy can be safely executed in an unseen scenario, we additionally include two constraints:

\begin{itemize}
    \item $\mathsf{IsReachable}(h,p)$, which is satisfied if a point $p$ is reachable for the robot gripper $h$.
    \item $\mathsf{SafeBiOp}(\mathfrak{a}, h_l, h_r, q_l, q_r)$, which is satisfied if the bimanual skill $\mathfrak{a}$ starting at the configuration $\langle q_l, q_r \rangle$ is safe from collision with any objects other than objects being manipulated. 
\end{itemize}

The implementation of $\mathsf{IsReachable}$ is straightforward, as it can be scripted using distance-related metrics. However, the $\mathsf{SafeBiOp}$ constraint for a learned reactive policy is not compatible with scripted test functions. 
The reason is that the trajectory \( \tau_\pi \) of a visuomotor policy $\pi$ cannot be explicitly computed, and the safety cannot be certified by a conventional collision checking function.

To predict the likelihood of collision during performing a coordinated skill $\mathfrak{a}$, we train a feasibility validator $\xi$, eliminating the need for an explicit trajectory. Specifically, an MLP  predicts a collision probability conditioned on the initiation conditions of the learned visuomotor policy.
Let the set of timestamps corresponding to $G_\mathrm{pre}, G_\mathrm{mid}, G_\mathrm{eff}$ be $T_\mathrm{pre} ,T_\mathrm{mid}, T_\mathrm{eff}$, respectively.  
Specifically,  we are interested in predicting if a bimanual operation starts from $t \in  T_\mathrm{pre}$ is collision-free during time $T_\mathfrak{a} = T_\mathrm{pre} \cap T_\mathrm{mid} \cap T_\mathrm{eff}$.
Denoting the unexpected obstacle with geometric feature $\nu$ as being located at $p$,   we construct a collision-prediction dataset as $ \bigcup ^{N-1}_{i=0} \underset{t\in T_\mathrm{pre}}{\bigcup} \underset{p}{\bigcup}  \{( \mathbf{q} ^i_t, p, \nu, \delta)\}$, where $\delta$ is the closest distance of the robot to the obstacle while performing skill $\mathfrak{a}$, and $N$ is the number of demonstrations. The predicate $\mathsf{SafeBiOp}$ is certified when all objects on the table are free from collision.

\section{Experiments and Results}
In this section, we evaluate SViP in both simulated and real-world tasks, to empirically answer following questions: 
\begin{itemize}
    \item \textbf{Q1}: Can SViP perform effectively in previously unseen tabletop setups? 
    \item \textbf{Q2}: Can SViP manage various constraints, such as reachability and collision avoidance?
    \item \textbf{Q3}: Can SViP complete tasks with novel goals using learned and scripted primitives?
\end{itemize}

\subsection{Simulation Experiment} \label{sec:sim}

\subsubsection{Setup of Simulated Environment} \label{sec:sim-env}
We adopt the peg-in-hole simulation environment from ALOHA \cite{zhao2024aloha} based on MuJoCo \cite{todorov2012mujoco} simulator. In this setup, two Interbotix vx300s robot arms are placed on the table with their $X$-axes oriented toward each other. A socket and a peg are placed on the left and right sides of the table, respectively. When collecting the training demonstrations, we employ the initial placement method \cite{zhao2024aloha} where the 2D position of each object is randomly sampled within a $0.2 m \times 0.2 m$ square, while their rotations remain fixed. We generate 50 episodes of demonstrations using a scripted motion planner, which is detailed in the open-source repository provided by \citet{zhao2024aloha}. The simulation experiments were computed on a workstation with Intel i7-13700k CPU and NVIDIA RTX 3090 GPU. 
\subsubsection{SViP Evaluation with OOD Observations}

In this part, we evaluate the performance of SViP when it encounters both in-distribution (ID) and out-of-distribution (OOD) observations during inference. To highlight the advancements in robustness and generalization offered by SViP, we conduct a comparative analysis with state-of-the-art (SOTA) generative IL methods, i.e., ACT \cite{zhao2023learning} and DP \cite{chi2023diffusion}.  

\paragraph{Design of Testing Scenario} \label{sec:sim_setup}
We design three types of randomized initialization functions to evaluate SViP's performance in unseen environments from a test distribution.
\begin{itemize}
    \item \textbf{\texttt{ID}} follows the object position sampling function used in demonstrations to generate in-distribution initial observations, ensuring alignment with the training dataset.
    \item \textbf{\texttt{XY-OOD}} randomly samples the 2D position of the object within $0.25 m \times 0.25 m$ square, which include the region defined as \texttt{ID}. This serves to create OOD observations limited to changes in the $XY$ plane.
    \item \textbf{\texttt{XYH-OOD}} introduces variability by rotating the object's heading within the range $\psi \in (-0.5\pi, 0.5\pi )$, in addition to the 2D position sampling in \texttt{XY-OOD}. This allows for a broader range of OOD scenarios with orientation changes.
\end{itemize}

\paragraph{Comparative Results and Analysis}
The task goal of SViP is defined as $\eta = \mathsf{DoneBiOp}(\mathfrak{a})$, indicating that the goal is to complete the bimanual insertion operation as demonstrated. The success rate is computed by averaging the results of 50 rollouts. 
The results are presented in Table~\ref{tab:sim_ood}. 

In the {\texttt{ID}} setting, a comparison of the success rates reveals that SViP surpasses both baseline methods.
The baseline performance is affected by the multi-stage nature of the peg-in-hole task, where the failure during the object pick-up phase can result in the failure of the entire task. In this context, using an object-centric motion primitive for object pickup proves to be more reliable than the IL policy, thereby enhancing the overall success rate.
In the {\texttt{XY-OOD}} setting, it is evident that ACT and DP experience significant performance degradation in OOD scenarios, while SViP maintains 100\% success.
In the {\texttt{XYH-OOD}} setting, the success rates of baselines drop dramatically, in contrast, SViP achieves 88\% success rate.
Therefore, SViP successfully executes bimanual multi-stage tasks in unseen tabletop setups, indicating its superior generalization to OOD initial scenarios (as in \textbf{Q1}).

\begin{table}
    \centering
        \caption{Success rates (\%) of SViP in the simulated \textit{Bimanual Insertion} task with different initial setups.}
    \begin{tabular}{c c c | c}
    \toprule
         \textbf{Setup} & \quad \textbf{ACT} \cite{zhao2023learning} & \textbf{DP} \cite{chi2023diffusion} \quad & \quad \textbf{SViP}  \\
     \midrule
       \texttt{ID}   & 44  & 54 & \textbf{100} \\
        \texttt{XY-OOD}  & 38 & 42 & \textbf{100}  \\
        \texttt{XYH-OOD} & 2 & 12 & \textbf{88}  \\
    \bottomrule
    \end{tabular}
    \label{tab:sim_ood}
    \vspace{-0.2in}
\end{table}

\begin{figure}[htb] 
\vspace{-1.5mm}
\subfloat[Unreachable Setup]
{\includegraphics[width=0.5\textwidth]{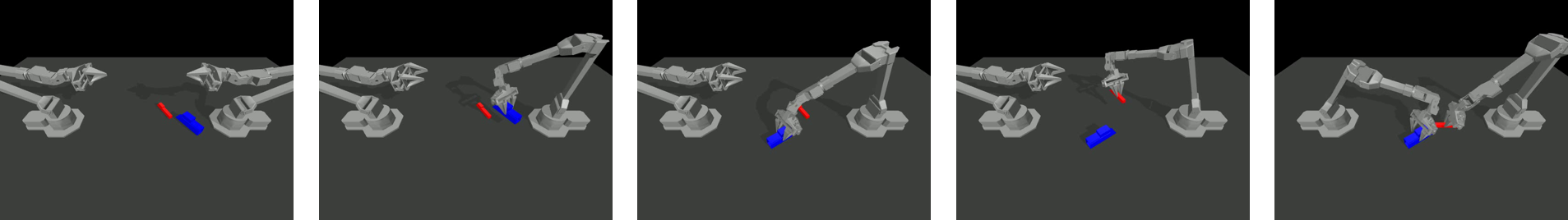} \label{fig:unreach_scene}
}\hfill 
\subfloat[Unsafe Setup]
{\includegraphics[width=0.5\textwidth]{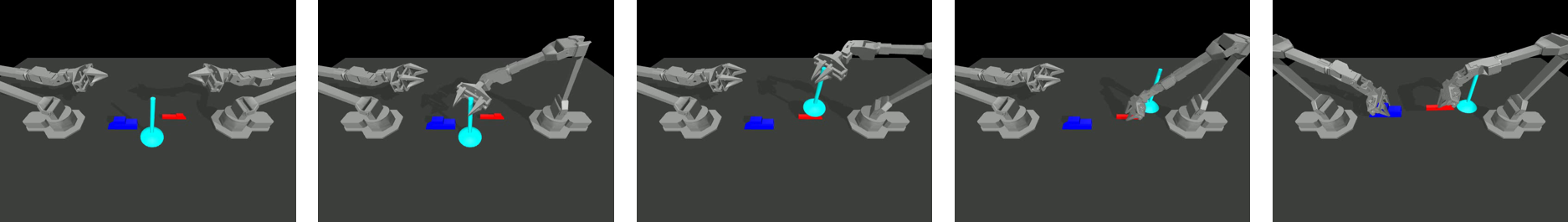}\label{fig:unsafe_scene}
}\hfill 
\caption{Execution by SViP in the \texttt{unreachable} and the \texttt{unsafe} setup.
}
\vspace{-3mm}
\end{figure}

\begin{table}
    \centering
    \caption{Results of SViP's zero-shot transfer to tasks that involve operations unseen in the demonstration. 
    }
    \begin{tabular}{p{1.7cm} p{1.7cm} p{1.7cm} p{1.7cm}}
    \toprule
         \textbf{Setup} &  \textbf{Sequence Len.} & \textbf{Success Rate} &  \textbf{Computation Time (s)}  \\
         \midrule
         \texttt{ID} (original) & 7.0  & 100 &  7.55 \\
       \texttt{unreachable}  & 11.0   & 55 & 12.36  \\
    \texttt{unsafe}  &  11.6 & 70 & 27.39  \\
    \bottomrule
    \end{tabular}
    \label{tab:sim_long}
    \vspace{-0.2in}
\end{table}

\subsubsection{SViP Evaluation in Constraint Handling}
To demonstrate the compatibility of SViP with the classical TAMP system, we evaluate SViP in two common environmental constraints: reachability constraints and collision-free constraints.

\paragraph{Design of Testing Scenario}
We have presented two important constraints, namely $\mathsf{IsReachable}$ and $\mathsf{SafeBiOp}$ in Section~\ref{sec feasibility}, to ensure the reliable execution of bimanual operations. To this end, we design two distinct simulation settings, named {\texttt{unreachable}} and {\texttt{unsafe}}, by updating the simulation environment setups.
Specifically, Fig.~\ref{fig:unreach_scene} depicts the {\texttt{unreachable}} scenario, where either the peg or the socket is inaccessible, preventing the insertion operation. In this case, both the socket and the peg are randomly placed on the same half of the table, with a random rotational angle $\psi \in (-\pi, \pi)$.
This setup effectively tests the system's adaptability to positioning constraints associated with reachability.
Fig.~\ref{fig:unsafe_scene} presents the {\texttt{unsafe}} scenario, where a pole is placed on the table, serving as a potential obstacle. Here, we assume that the pole can be grasped and its base is not fixed. In this setting, the placement of the socket and the peg adheres to the \texttt{ID} configuration as described in Section ~\ref{sec:sim_setup}. This scenario is designed to examine the system's capacity to handle the collision-free constraints effectively.

\paragraph{Results and Analysis}
We conducted 20 trials for each task using SViP on a workstation with Intel i7-13700k CPU and NVIDIA RTX 3090 GPU.  Several key metrics are recorded: the average number of actions (i.e., sequence length), the success rate (\%), and the average computation time (s).  We set a timeout threshold of 60 seconds, and any trial exceeding this limit is deemed a failure
It is important to note that both baseline methods, ACT\cite{zhao2023learning} and DP\cite{chi2023diffusion}, completely failed in \texttt{unreachable} or \texttt{unsafe} scenarios. For the sake of simplicity and clarity, their results have been excluded from the table. 
As indicated in Table~\ref{tab:sim_long}, SViP achieves success rates of 55\% and 70\% in the unreachable and unsafe scenarios, respectively. 
This performance is primarily attributed to the integration of TAMP.
TAMP can identify inadequacies in the demonstrated sequence, thereby replanning a longer plan skeleton that achieves the goal conditions without violating constraints. 
For example, in the {\texttt{unreachable}} scenario, the robot can use its left arm to grasp the socket that is initially out of reach, and place it into the right arm's reachable region. Similarly, in the {\texttt{unsafe}} scenario, the robot may first relocate the pole to ensure it does not obstruct further insertion operations. 
Despite these capabilities, Table~\ref{tab:sim_long} indicates a reduction in the success rate and an increase in computation time compared to the results in the \texttt{ID} setup.
Here, the primary cause of decreased success rates is unexpected collisions, which stem from inaccurate point cloud observations. Regarding the increased computation time, especially in the {\texttt{unsafe}} case, this is due to the exponential rise in computational complexity of TAMP as the number of objects increases. 
Nevertheless, unlike IL policies, which cannot handle scenarios such as \texttt{unreachable} and \texttt{unsafe}, SViP represents a significant advancement by incorporating reachability and collision-avoidance constraints, addressing the primary concern of \textbf{Q2}. 

\subsection{Real-world Experiment}

\subsubsection{Real-world Tasks and Robot Platform}
\begin{figure}[htb] 
\vspace{-2mm}
\subfloat[Screwdriver Packing]
{\includegraphics[width=0.5\textwidth]{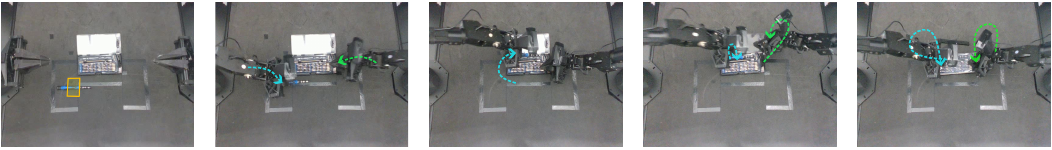}  \label{fig:screw-seq}
\vspace{-2mm}
}\hfill 
\subfloat[Cup-sleeve Insertion]
{\includegraphics[width=0.5\textwidth]{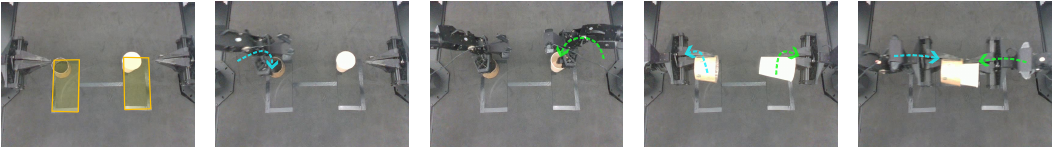}  \label{fig:cup-seq}
}\hfill 
\caption{Demonstrated manipulation in the \textit{Screwdriver Packing} and the \textit{Cup-sleeve Insertion} task. The initial positions of the manipulated objects are randomly sampled in the yellow regions.
}
 \vspace{-0.1in}
\end{figure}

This section evaluates SViP in three real-world tasks: 1) \textit{Object Handoff}, as previously introduced in Section~\ref{sec:components}; 2) \textit{Screwdriver Packing}, which requires fine-grained manipulation of a slender object; and 3) \textit{Cup-sleeve Insertion}, involving interaction between two objects. 
Fig.~\ref{fig:screw-seq} and Fig.~\ref{fig:cup-seq} illustrate the demonstrated manipulation for  \textit{Screwdriver Packing} and \textit{Cup-sleeve Insertion} tasks, respectively. For each task, we collected 20 demonstrations for training the visuomotor policy. 
We conducted the experiments on ALOHA \cite{zhao2023learning} platform, consisting of two master arms and two puppet arms, each with 7-DoF. The master arms are only used for teleoperation, and the puppet arms precisely mirror the joint positions of the master arms. Notably, only the puppet arms are activated during execution. 
The ALOHA system is equipped with four cameras. One top camera and one front camera capture the overall workstation, along with two cameras mounted on the wrists to record the detailed visual information. Note that we set the top camera as an Intel Realsense camera, which provides depth measurements for 3D perception.  

\subsubsection{SViP Evaluation in Real-World Tasks}
We design the experiments with three different setups: in-distribution (\texttt{ID}), out-of-distribution (\texttt{OOD}), and \texttt{Tilted}. 
Notably, the \texttt{Tilted} setup presents a more challenging scenario for the \textit{Object Handoff} and \textit{Screwdriver Packing} tasks, as it involves spatial rotations where the target object leans against a random object. It is important to highlight that the objects involved in the \textit{Object Handoff} task and \textit{Cup-sleeve Insertion} tasks exhibit central symmetry, posing challenges for most object pose estimators \cite{liu2022gen6d,wen2024foundationpose} that fail to provide stable pose outputs. 
Additionally, we choose DP as the sole baseline due to its superior performance compared to ACT, as illustrated in Section~\ref{sec:sim}).
For each task, we conducted 20 rollouts and calculated the success rate for each setup. As shown in Table~\ref{tab:real_ood}, SViP consistently outperforms DP across all task scenarios. Similar to the simulation results, SViP demonstrates superior performance in \texttt{ID} scenarios, even with only 20 real-world demonstrations, highlighting its data efficiency and reliability.
Notably, in the fine-grained, multi-stage manipulation task \textit{Screwdriver Packing}, even minor OOD observations cause DP to fail the entire task. In contrast, SViP maintains high success rates across all settings.
Despite its overall effectiveness, the accuracy of switching condition generators is affected by OOD point cloud observations, particularly in the \texttt{Tilted} setting of the \textit{Cup-sleeve Insertion} task. We attribute this issue to the reliance on single-view depth measurements, which inherently produce noisy and incomplete point clouds, leading to discrepancies between training and inference observations. 

\begin{table}
    \centering
    \caption{Success rate ($\%$) of SViP in real-world tasks with various initial setups, in comparison with DP.}
    \vspace{-0.1in}
    \begin{tabular}{c c | c c }
    \toprule
       \textbf{Task} & \textbf{Setup} \quad & \quad \textbf{DP} \cite{chi2023diffusion} & \textbf{SViP} \\ \midrule
\multirow{5}{*}{Object Handoff} & ID &  {90} &  \textbf{100} \\ \cmidrule(lr){2-4}
& OOD & 30  &\textbf{95} \\  \cmidrule(lr){2-4}
& Tilted & 60 & \textbf{100} \\ \midrule
\multirow{5}{*}{Screwdriver Packing} & ID & 55 &  \textbf{80} \\ \cmidrule(lr){2-4}
& OOD & 10 & \textbf{80} \\ \cmidrule(lr){2-4}
& Tilted & 0 & \textbf{65} \\ \midrule
\multirow{5}{*}{Cup-sleeve Insertion} & ID & 75 & \textbf{80}  \\ \cmidrule(lr){2-4}
& OOD & 35 & \textbf{75} \\ \cmidrule(lr){2-4}
& Tilted & 35 & \textbf{40} \\
    \bottomrule 
    \end{tabular}
    \label{tab:real_ood}
    \vspace{-0.2in}
\end{table}

\subsubsection{SViP Evaluation in Novel Tasks with Specified Goal}

The experiments in this section are designed to demonstrate that the learned bimanual visuomotor policy is capable of executing long-horizon tasks that align with a human-specified goal. It is important to know that if the task goal contains predicates regarding the state of interested objects, the effect of the bimanual operation should be well-defined.
For illustration, we design two specific tasks as follows:
\begin{itemize}
    \item \textit{Table-to-Bin Clearance}: Based on the \textit{Object Handoff} scenario, we additionally place a bin at the top-left corner of the table. The task goal is defined as $\eta = \underset{o_i \in O}{\bigcap} \mathsf{In} \text{(bin},o_i)$, indicating that all objects on the table must be within the bin. 
    \item \textit{Cup-sleeve Insertion and Reconfiguration}: Based on the \textit{Cup-sleeve Insertion} scenario, we add two virtual pads, namely \textit{leftPad} and \textit{rightPad}, on the table.
    An additional goal condition is defined as $\mathsf{On} (\text{sleeve, leftPad)}) \wedge \mathsf{On}( \text{cup, rightPad})$, requiring the robot to place the cup and sleeve to the table at arbitrary positions. 
    The geometric parameters remain consistent with those in \textit{Cup-sleeve Insertion (OOD)} scenario.
    \item \textit{Multiple Instructions}: The goal is given as $\eta = \underset{a_i \in \{a_0, a_1\}}{\bigcap}\mathsf{DoneBiOp}(\mathfrak{a_i}) \wedge \underset{o_i \in O}{\bigcap} \mathsf{On}(o_i, \text{table})$, where $a_0$ and $a_1$ are the \textit{screwdriver-packing} and the \textit{cup-cleaning} actions, respectively. Apart from  $a_0$ and $a_1$, another skill, handing off the cup, is incorporated as $a_2$ for task completion.
\end{itemize}

\begin{figure}[t] 
\vspace{1.5mm}
\subfloat[Table-to-Bin Clearance]
{\includegraphics[width=0.5\textwidth]{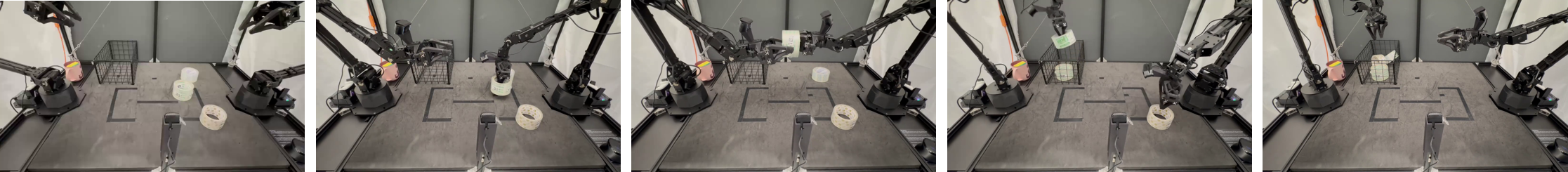}  \label{fig:3tape}
\vspace{-2mm}
}\hfill 
\subfloat[Multiple Instructions]
{\includegraphics[width=0.5\textwidth]{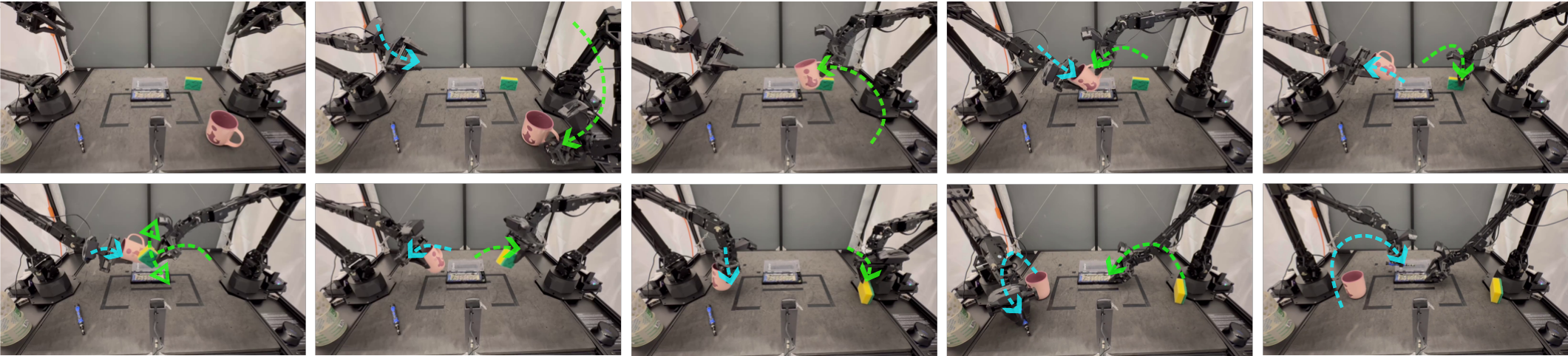}  \label{fig:screw-hand-wipe}
}\hfill 
\caption{The screenshots of rollouts in (a) \textit{Table-to-Bin Clearance} task and (b) \textit{Multiple Instructions} task.
}
\vspace{-6mm}
\end{figure}

The execution histories of \textit{Table-to-Bin Clearance} task and \textit{Multiple Instructions} task can be viewed in Fig.~\ref{fig:3tape} and Fig.~\ref{fig:screw-hand-wipe}, respectively. Particularly, in \textit{Multiple Instructions} task, SViP sequences multiple learned bimanual visuomotor policies to complete a long-horizon manipulation containing 19 discrete steps. The robot first hands off the cup from right to left, then grabs a sponge to wipe the cup, and finally puts down objects and packs the screwdriver. For the \textit{Cup-sleeve Insertion and Reconfiguration} task,  the bimanual insertion and object placement are conducted in a repetitive manner,  thereby saving time for manual reconfiguration of the tabletop. The results demonstrate that SViP effectively tackles novel tasks by leveraging compositionally learned skills, addressing the concerns in \textbf{Q3}.

\section{Conclusion} \label{sec:conclusion}
We proposed SViP, a data-efficient system that leverages the advantages of both visuomotor policies and long-horizon planning for generalized robot tasks. SViP identifies the contact-rich segments in a bimanual manipulation as a black-box operation, and extracts contact modes as symbolic skill descriptions.
With the help of the proposed switching condition generator, object-centric unimanual operations are smoothly stitched with the learned visuomotor policy.
SViP demonstrates its ability to handle out-of-distribution initial observations, accomplish tasks unattainable through teleoperation alone, and achieve novel goals by automatically composing learned skills.

Nevertheless, we admit certain limitations in this work. First, TAMP requires manual PDDL definitions, although the effort is reduced by the automatic predicate parsing from scene graphs. Second, the partial point cloud observation from a single-view depth measurement limits the SViP's performance. In the future, we plan to equip SViP with online reconstruction capability for more accurate 3D perception. 

\bibliographystyle{plainnat}

%
%
\end{document}